\RequirePackage{fix-cm}
\documentclass[twocolumn]{svjour3}     
\smartqed 
\usepackage{graphicx}
\usepackage{natbib}

\usepackage{soul}
\usepackage{url}
\usepackage[utf8]{inputenc}
\usepackage{graphicx}
\usepackage{amsmath}
\usepackage{booktabs}
\usepackage{algorithm}
\usepackage{algorithmic}
\usepackage{epsfig}
\usepackage{amssymb}
\usepackage{color}
\usepackage[table]{xcolor}
\usepackage{hhline}
\usepackage{stfloats}
\definecolor{Gray}{gray}{0.85}
\usepackage{diagbox}
\usepackage{multirow}

\newcommand{\pluseq}{\mathrel{+}=}

\newcommand{\eqnref}[1]{Eq.~(\ref{eqn:#1})}
\newcommand{\figref}[1]{Fig.~\ref{fig:#1}}
\newcommand{\tabref}[1]{Table~\ref{tbl:#1}}
\newcommand{\secref}[1]{Section~\ref{sec:#1}}

\usepackage{dsfont}
\usepackage{pifont} 

\usepackage{float} 

\usepackage{hyperref}

\newcommand{\DissectionPreview}{
\begin{figure*}[t]
    \centering
    \includegraphics[width=.9\linewidth]{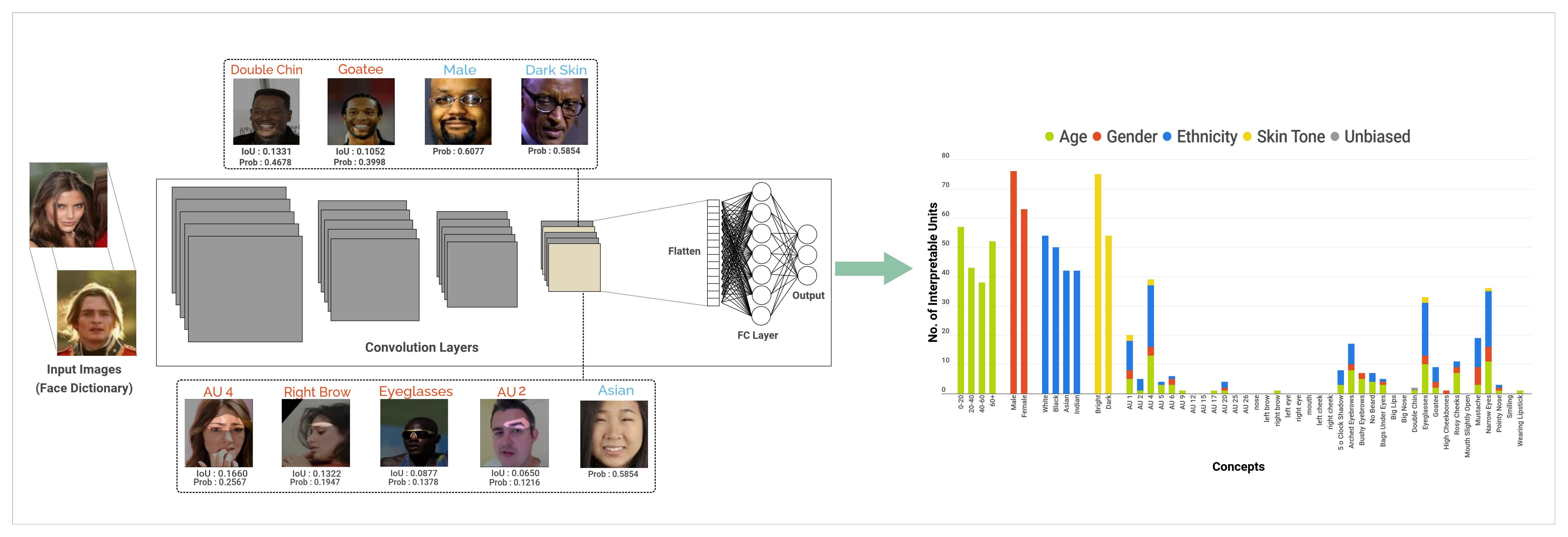}
    \caption{Overview of Hierarchical Network Dissection for face-centric CNNs. Given a trained model, Hierarchical Network Dissection detects units that act as facial concept detectors and pairs the units with the corresponding concept. The two boxes of the left side of the figure display samples of concepts paired with the respective units, where the local concepts (Red) have an IoU score and a probability, whereas global concepts (Blue) only have a probability score. The right side of the figure shows a dissection report, displaying the histogram of all the concepts that have been paired with units.}
    \label{fig:dissection_preview}
\end{figure*}
}

\newcommand{\figDictionary}{
\begin{figure*}[t]
    \centering
    \includegraphics[width= \linewidth]{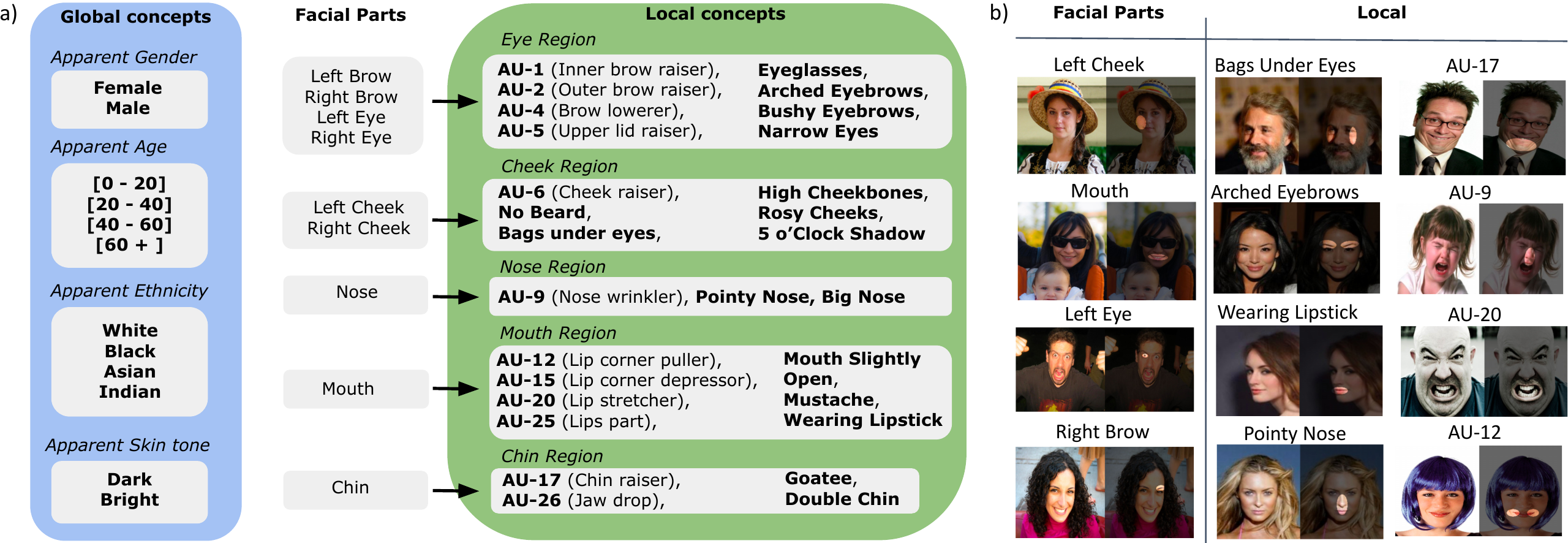}
    \caption{Face Dictionary: a) Concepts included in our face dictionary, organized by type (Global \& Local). Notice that the Local concepts are grouped by Facial Parts; b) Examples of images and corresponding segmentation masks for facial parts and for each type of Local Concept (i.e. facial attributes and Action Units).}
        \label{fig:fig_dictionary}
\end{figure*}
}

\newcommand{\DictionaryHist}{
\begin{figure*}[t!]
    \centering
    \includegraphics[width=\linewidth]{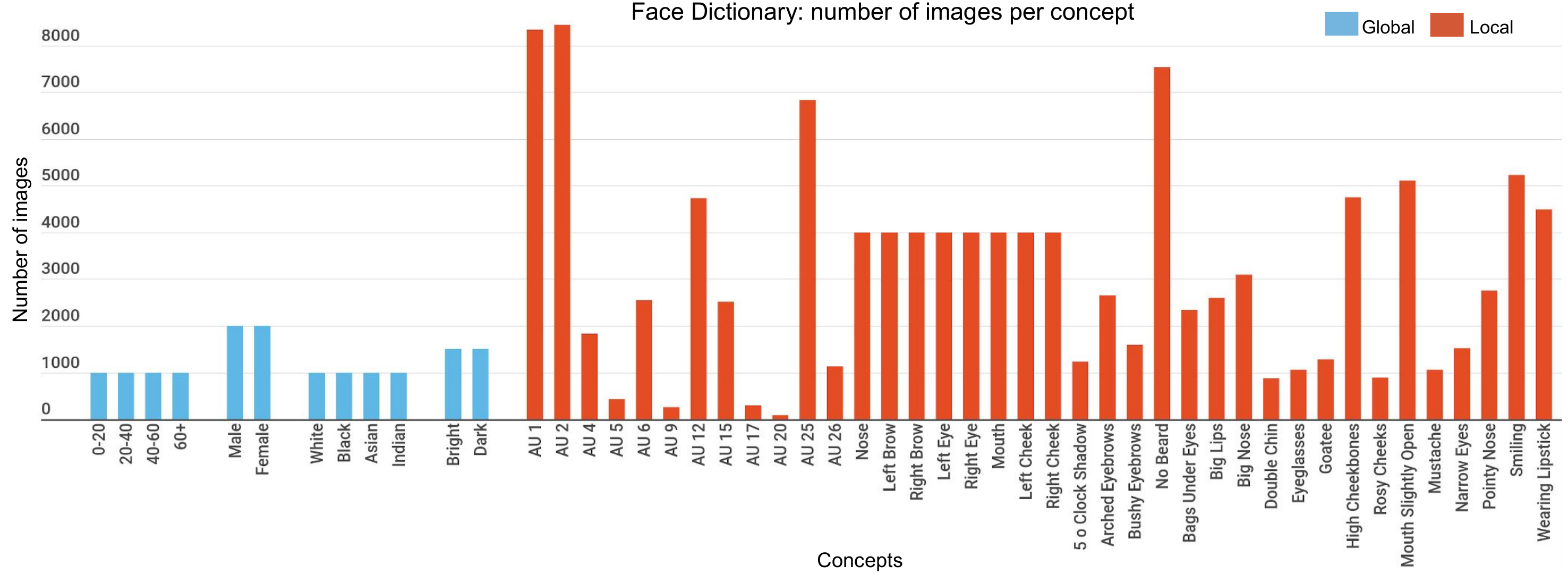}
    \caption{Number of images per concept in our Face Dictionary.}
    \label{fig:dictionary_breakup}
\end{figure*}
}

\newcommand{\ProbViz}{
 \begin{figure*}[t!]
  	\centering
  	\includegraphics[width=\textwidth]{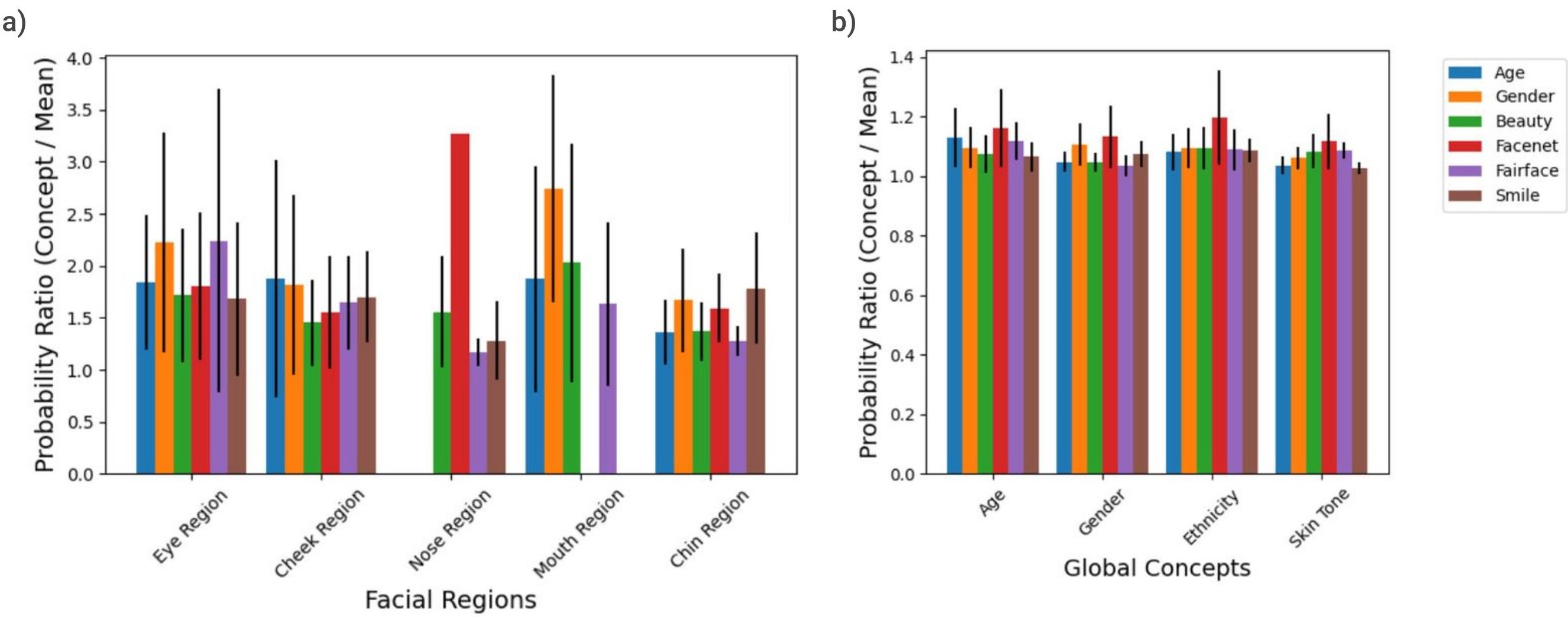}
    \caption{Mean probability visualization for each model: a) per Facial Region; b) per Global Concept. These probabilities are only counted when they are above the mean probability of their region.}
    \label{fig:ProbViz}
    \vspace{-.30cm}
\end{figure*}
}

\newcommand{\dissectingfacemodels}{
\begin{figure}[b]
    \centering
    \includegraphics[width=\linewidth]{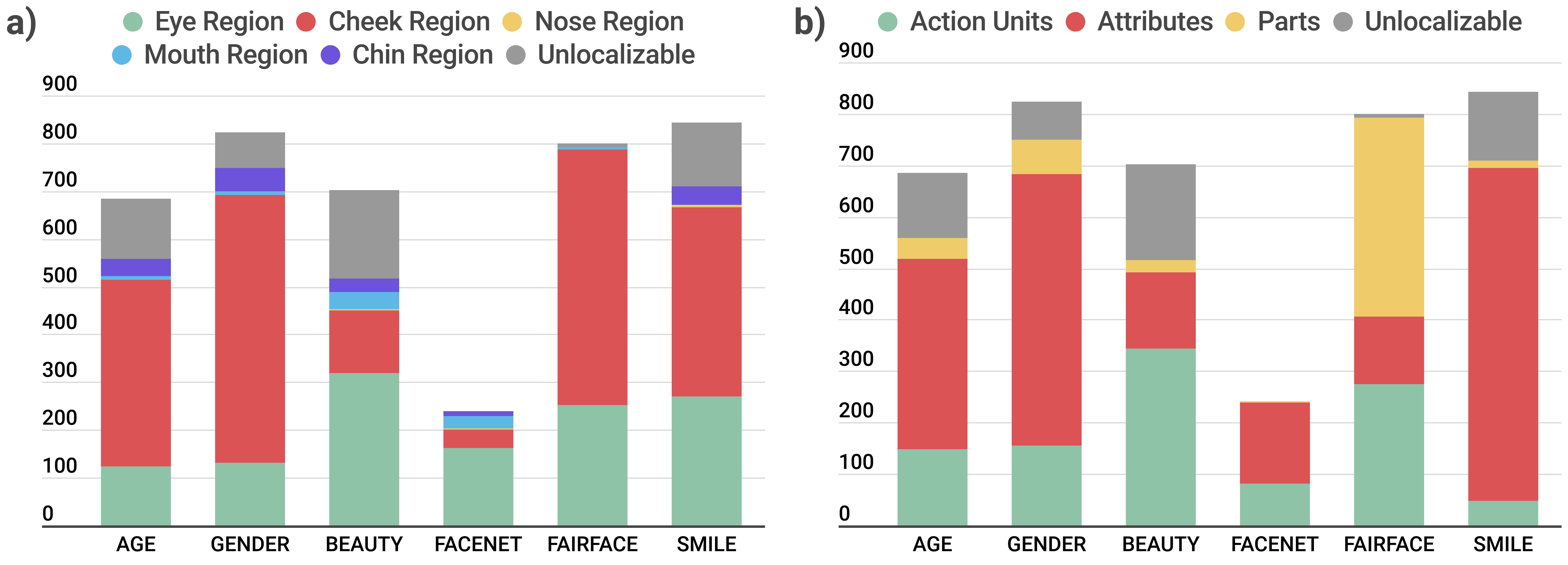}
    \caption{Per each model: a) Distribution of interpretable units per face part; b) Distribution of types of concepts.}
    \label{fig:dissecting_face_models}
      \vspace{-.5cm}
\end{figure}
}

\newcommand{\HDissResults}{
\begin{figure*}[ht!]
    \centering
    \includegraphics[width=\linewidth]{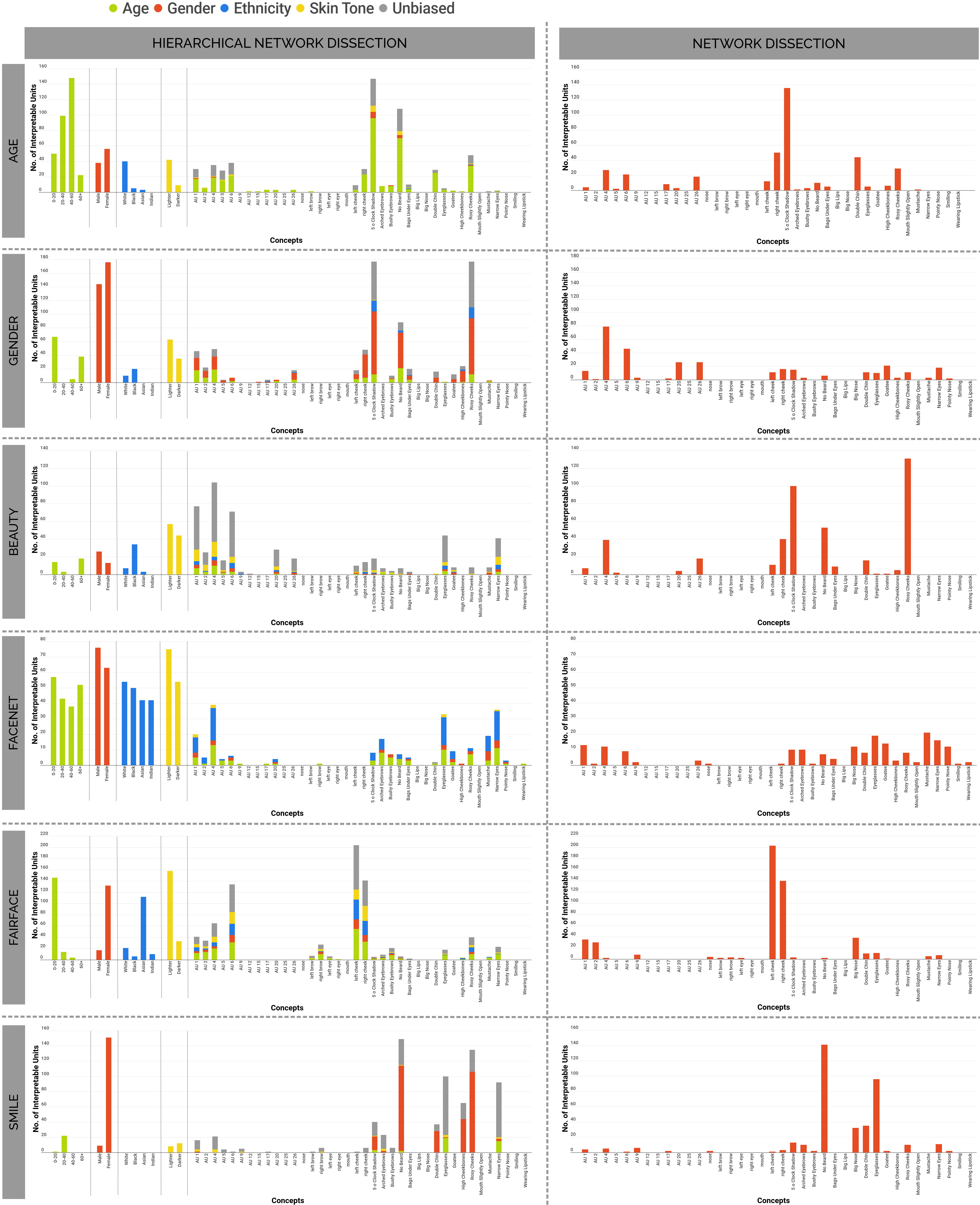}
    \caption{Complete dissection histograms of the six models (listed in \tabref{CNNmodel}) showing the number of interpretable units for both global and local concepts. Number of units paired with each global concept subgroup is displayed alongside the local concepts. Local concepts are visualized by mapping them with global concepts, e.g. each column displays the number of interpretable units for that local concept that are also paired with global concepts in our dictionary.}
        \label{fig:NewDissectionResults}
\end{figure*}
}

\newcommand{\ActivationResults}{
\begin{figure*}[t!]
    \centering
    \includegraphics[width= \linewidth]{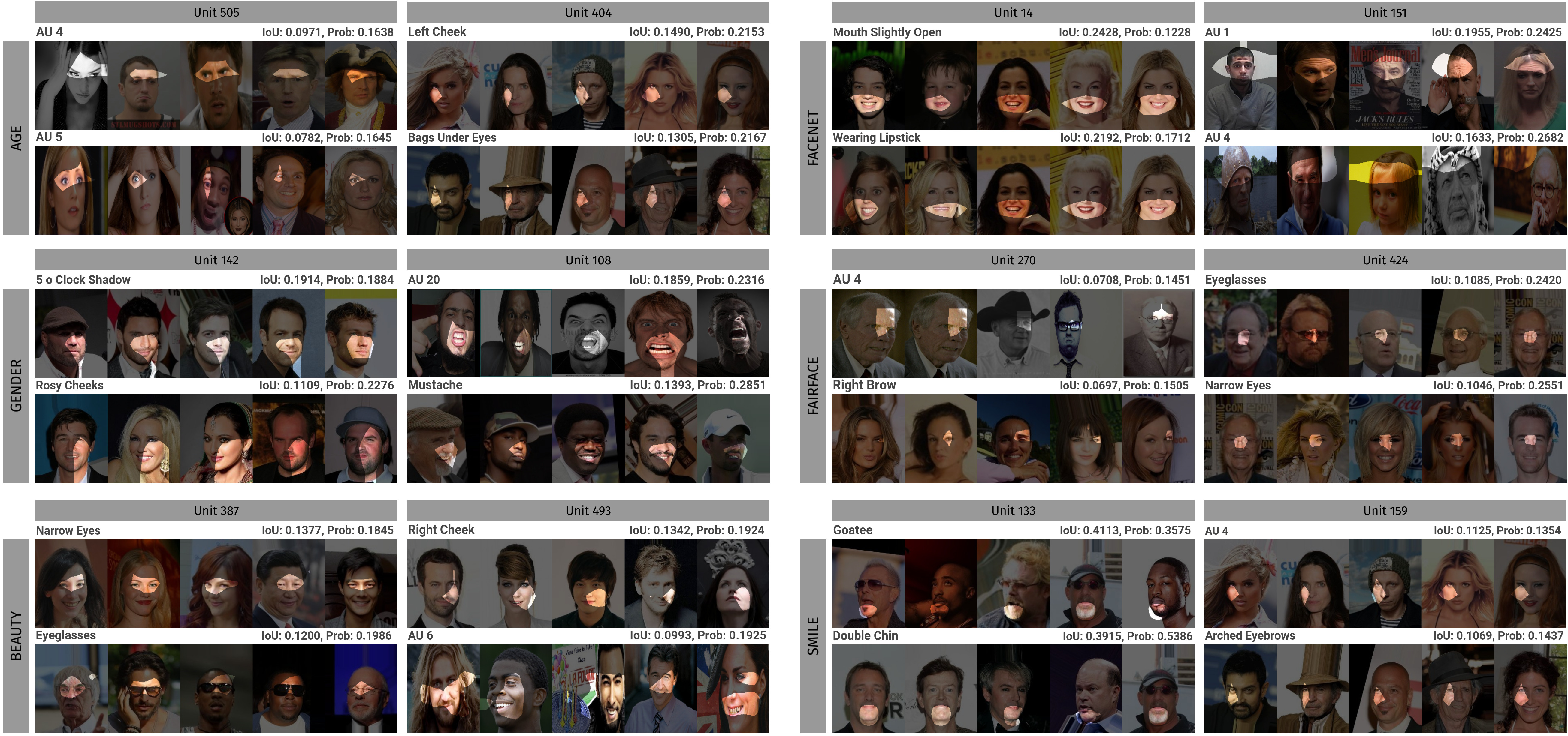}
    \caption{For all $5$ tasks, 2 examples of different concepts identified by Network Dissection(top) and Hierarchical Network Dissection(bottom). Our formulation identifies concepts ignored by the original formulation and obtains a higher probability score than the concept with a better IoU. The 5 images displayed have the highest IoU for that concept respectively.}
        \label{fig:ActivationResults}
\end{figure*}
}

\newcommand{\figCoverageComb}{
\begin{figure*}[ht!]
    \centering
    \includegraphics[width=0.95\linewidth]{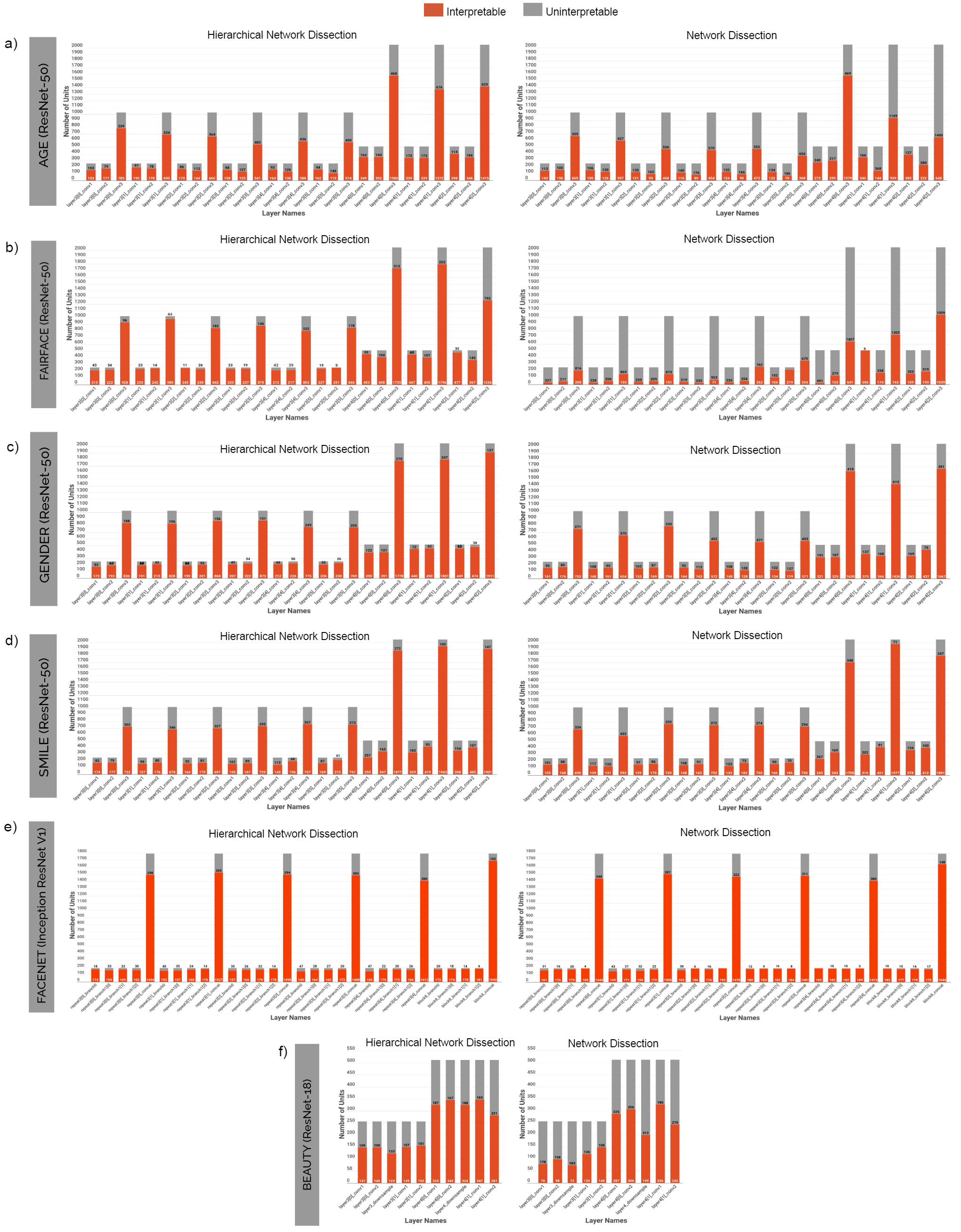}
    \caption{Number of interpretable units within multiple layers lying in the deeper stages of the networks per task. The layer names have the stage of the network (Layer3 / Layer4) followed by the sub-block number in square brackets and then the convolution number.}
        \label{fig_coverage_comb}
\end{figure*}
}

\newcommand{\Simulatedglassesimages}{
\begin{figure}[t]
    \centering
    \includegraphics[width=\linewidth]{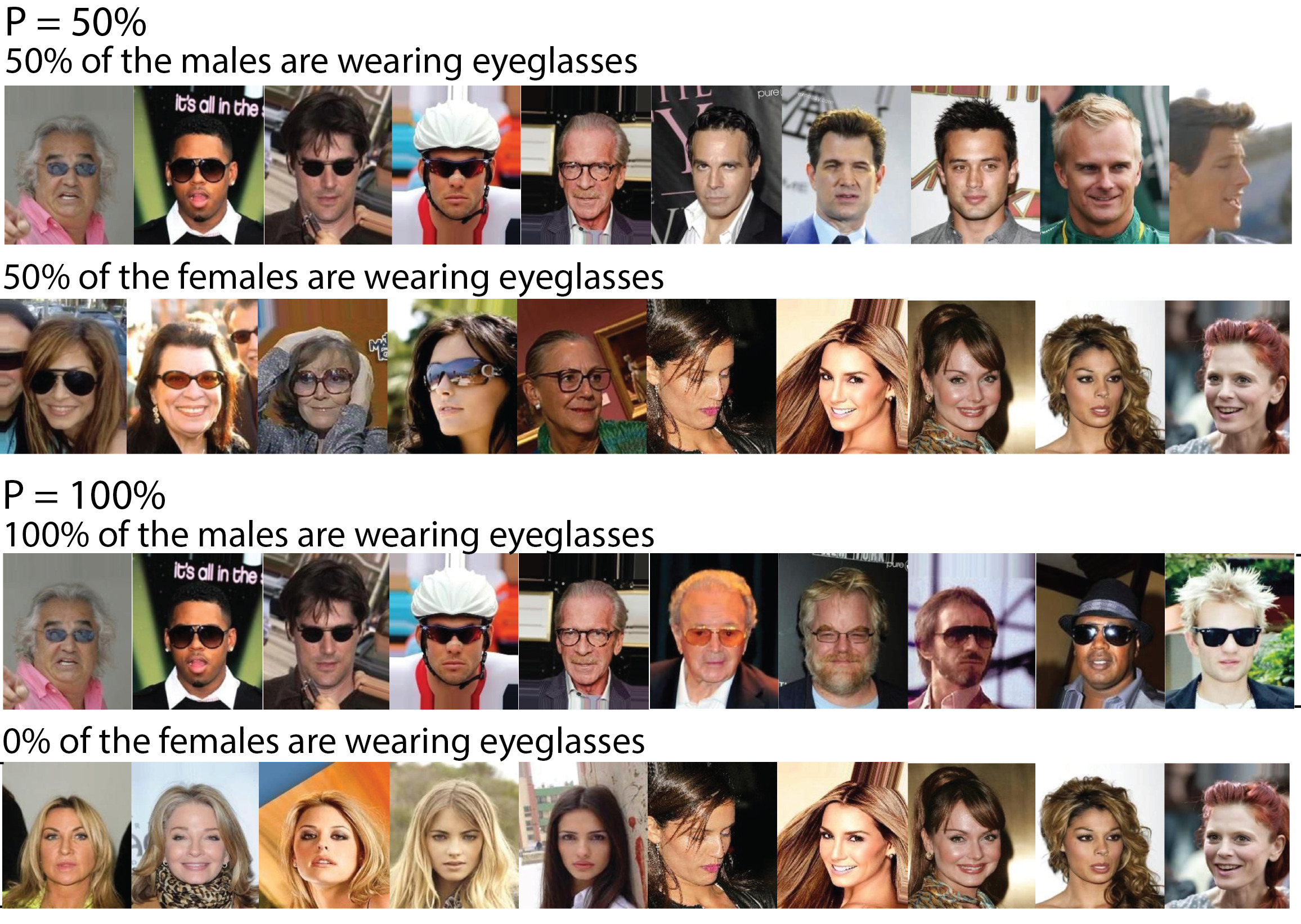}
    \caption{Illustration of bias introduced through ``Eyeglasses'' in a gender classification model with extreme probability $P$ values.}
    \label{fig:eyeglasses_bias}
\end{figure}
}

\newcommand{\TopGlassesImages}{
\begin{figure*}[t!]
    \centering
    \includegraphics[width=\linewidth]{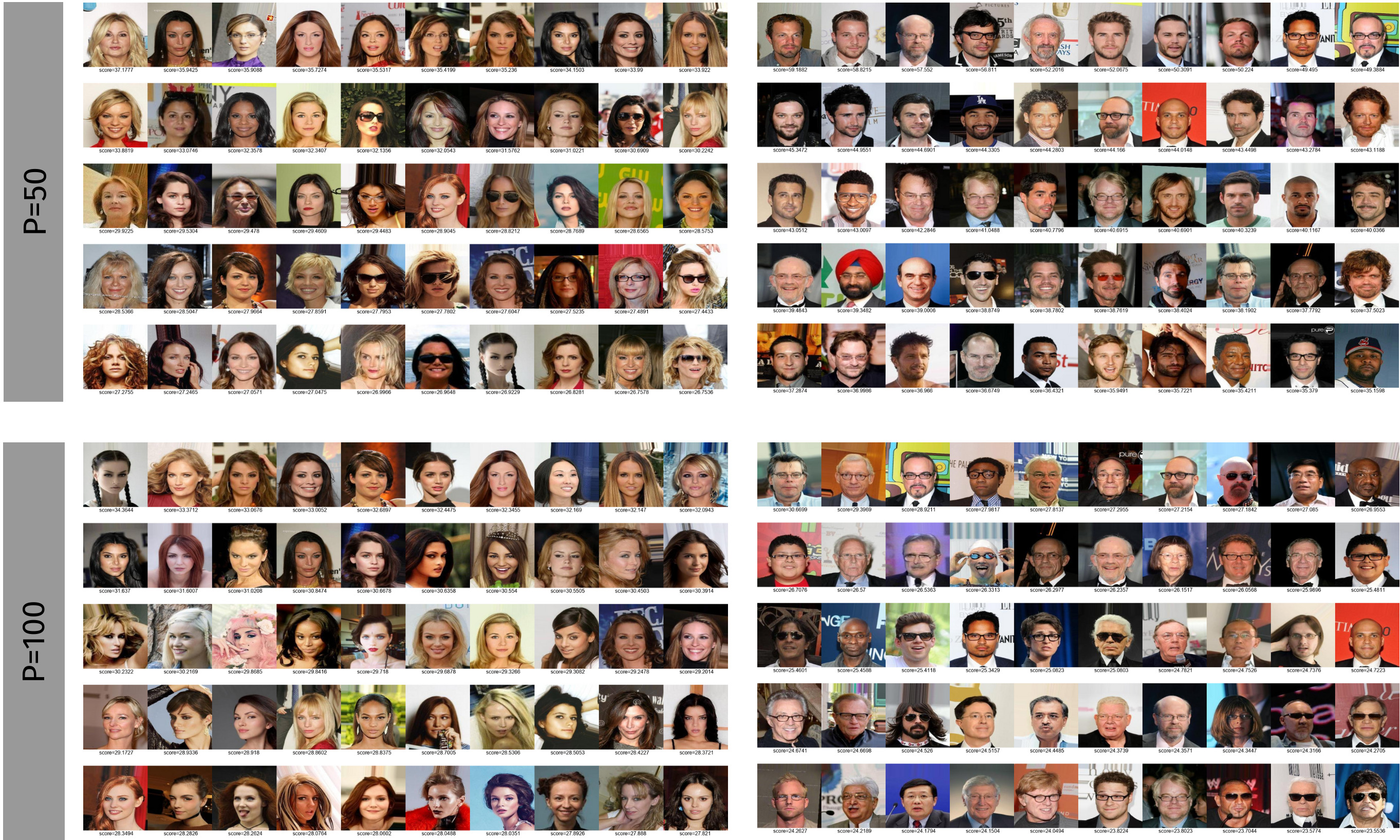}
    \caption{Top 50 images with the highest activation score per gender for models with P=50 and P=100 of ``Eyeglasses'' experiment.}
    \label{fig:top_gender_glasses}
\end{figure*}
}

\newcommand{\SimulatedExp}{
\begin{figure}[ht]
    \centering
    \includegraphics[width=\linewidth]{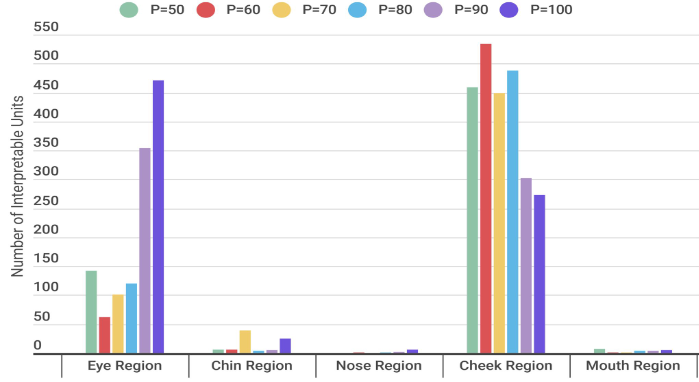}
    \caption{Face region concepts found in the models trained with bias on the ``Eyeglasses'' concept.}
        \label{fig:results_eyeglasses_bias}
\end{figure}
}

\newcommand{\grayscalePROBS}{
\begin{figure}[ht]
    \centering
    \includegraphics[width=\linewidth]{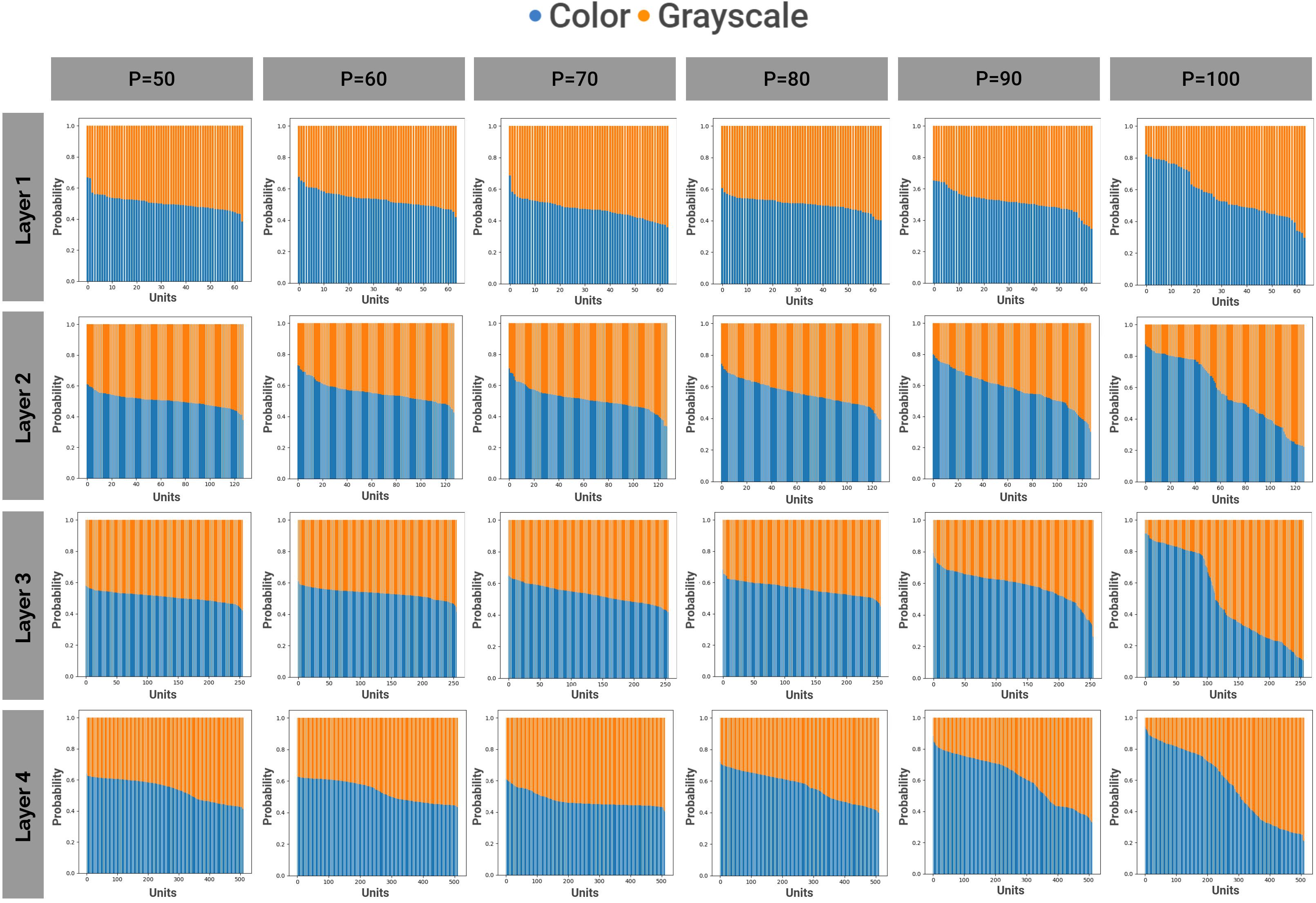}
    \caption{Unit Probabilities estimated using Stage I approach for ``Grayscale'' and ``Color'' schemes across all 4 layers in the models trained with bias on the color scheme. The units are arranged in descending order of color probability to visualize the scale of bias across units within a layer.}
    \label{fig:grayscale_probs}
\end{figure}
}

\newcommand{\nonlocalizable}{
\begin{figure}[ht]
    \centering
    \includegraphics[width=\linewidth]{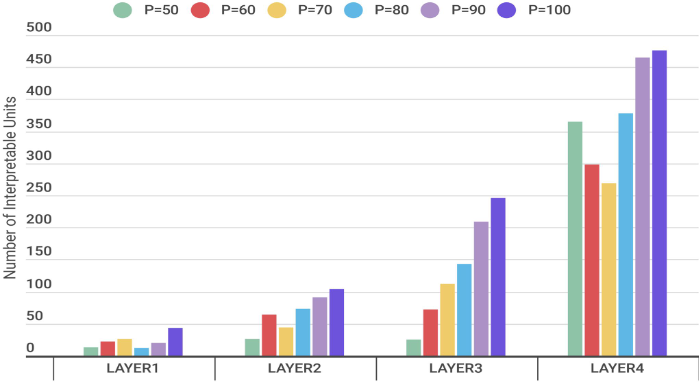}
    \caption{Number of biased units found across all 4 layers in the models trained with bias on the color scheme.}
    \label{fig:non_localizable_results}
\end{figure}
}

\newcommand{\tblCNNmodelnew}{
\begin{table*}[t]
\caption{Details of dissected face-centric CNN models. The metrics in the last column are Mean Absolute Error (MAE), Accuracy (Acc.), and  Mean Squared Error (MSE), respectively.}
\centering
\tiny
\begin{tabular}{|c | c | c | c| c |}
\hline
\textbf{Task (Abbreviation)}  & \textbf{Dataset}  & \textbf{Architecture} & \textbf{Layer Dissected}  & \textbf{Performance (Metric)} \\
\hline
Age Estimation (AGE) & IMDB-WIKI & ResNet-50 & Block 4 - Layer2(conv1) & 6.64 (MAE) \\
\hline                                       
Gender Classification (GENDER)& IMDB-WIKI   & ResNet-50 & Block 4 - Layer2(conv1) & 90.8\% (Acc.)                     \\
\hline
Beauty Estimation (BEAUTY)  & SCUT-FBP5500 & ResNet-18 (pretrained) & Block 4 - Layer1(conv1) & 0.137 (MSE)            \\
\hline
Facial Recognition (FACENET) &  VGGface2 & Inception ResNetV1 (pretrained) & Block 8 - branch1(conv3) &  99.6\% (Acc.)                   \\
\hline
Facial Recognition (FAIRFACE)   & FairFace & ResNet-50 & Block 4 - Layer2(conv1)  & 86.8\% (Acc.)                     \\
\hline
Smile Classification (SMILE) & CelebA & ResNet-50 & Block4 - Layer2(conv1) & 91.2 (Acc.) \\
\hline
\end{tabular}
\label{tbl:CNNmodel}
\end{table*}
}

\begin{document}
\sloppy
\title{Interpreting Face Inference Models using Hierarchical Network Dissection}


\author{Divyang Teotia$^{\ast}$  \and Agata Lapedriza$^{\mathsection}$ \and
    Sarah Ostadabbas$^{\ast}$  
}


\institute{$^{\ast}$ D. Teotia and S. Ostadabbas \at
       Augmented Cognition Lab, Electrical and Computer Engineering Department, Northeastern University.\\
       $^{\mathsection}$ A. Lapedriza \at Universitat Oberta de Catalunya.
}

\date{Accepted: March 2022}

\maketitle
\begin{abstract}
This paper presents Hierarchical Network Dissection, a general pipeline to interpret the internal representation of face-centric inference models. Using a probabilistic formulation, our pipeline pairs units of the model with concepts in our ``Face Dictionary'', a collection of facial concepts with corresponding sample images. Our pipeline is inspired by Network Dissection, a popular interpretability model for object-centric and scene-centric models. However, our formulation allows to deal with two important challenges of face-centric models that Network Dissection cannot address: (1) spacial overlap of concepts: there are different facial concepts that simultaneously occur in the same region of the image, like ``nose'' (facial part) and ``pointy nose'' (facial attribute); and (2) global concepts: there are units with affinity to concepts that do not refer to specific locations of the face (e.g. apparent age). We use Hierarchical Network Dissection to dissect different face-centric inference models trained on widely-used facial datasets. The results show models trained for different tasks learned different internal representations. Furthermore, the interpretability results can reveal some biases in the training data and some interesting characteristics of the face-centric inference tasks. Finally, we conduct controlled experiments on biased data to showcase the potential of Hierarchical Network Dissection for bias discovery. The results illustrate how Hierarchical Network Dissection can be used to discover and quantify bias in the training data that is also encoded in the model. \footnote{The ``Face Dictionary'' and pre-trained models are available at: \href{https://web.northeastern.edu/ostadabbas/2021/01/11/quantifying-notion-of-bias-in-face-inference-models/}{Hierarchical Network Dissection Webpage}. The source code is available at: \href{https://github.com/ostadabbas/Hierarchical-Network-Dissection}{Hierarchical Network Dissection GitHub.}}.
\keywords{Bias Discovery \and Face-based Inference Models \and Face Dictionary \and Model Interpretability \and Network Dissection.}

\end{abstract}

\section{Introduction}
\label{sec:intro}

Over the years, significant improvements have been made in developing deep learning models for face recognition \cite{parkhi2015deep}, apparent age and gender estimation \cite{eidinger2014age}, and facial attribute classification \cite{liu2019understanding}, among many others. However, when these face-centric models are applied in commercial settings at scale, they have been scrutinized for displaying biases towards underrepresented classes \cite{buolamwini2017gender,buolamwini2018gender}. 
These models have been shown to be flawed and inaccurate, specially in certain groups of population, such as dark skin tone faces.

Understanding the underlying reasons for these biases, which is necessary for making progress towards bias mitigation strategies, has been hindered since little to no work has been done to investigate the representations learned by these face-centric models. Studying the representations  learned by deep learning architectures has been mainly addressed for object-centric and scene-centric models \cite{zeiler2014visualizing,simonyan2014deep, mahendran2015understanding,network_dissection_cvpr, nguyen2016synthesizing}, showing how the model interpretation is useful for understanding the outputs of the network and, more generally, the behaviour of the trained model \cite{network_dissection_pami}.



\DissectionPreview
In this paper, we present Hierarchical Network Dissection, a general pipeline to interpret face inference models. Concretely, our pipeline uses a probabilistic formulation to reveal what units of a face-centric inference model act as detectors of facial concepts. As a result, the output of Hierarchical Network Dissection is a "dissection report", i.e. a histogram of the concepts learned by the model, as illustrated in Fig. 1.
Hierarchical Network Dissection uses our visual dictionary of facial concepts, called ``Face Dictionary'', which includes global concepts (like apparent age or apparent gender), and local concepts, which include action units and facial attributes. 
Our Face Dictionary is described in in \secref{dictionary}.

Our formulation is inspired by the Network Dissection approach proposed by \cite{network_dissection_cvpr, network_dissection_pami}, which is based on the observation that some units in scene-centric models act as object detectors \cite{object_detectors_emerge}. In fact, our pipeline for dissecting face-centric models uses Network Dissection in one of its three stages. However, there are important technical differences between the two formulations, since interpreting face-centric models presents new challenges compared to interpreting scene-centric models. First, Network Dissection is based on the Broadly and Densely Labeled Dataset (Broden) visual dictionary, which is a collection of scene-centric visual concepts, like colors, textures, materials, parts of objects, whole objects, and scenes. Notice that, while one can expect to find detectors of these concepts in scene-centric images, these concepts are barely relevant to the faces. Thus, dissecting face-centric models requires a completely new face-centric dictionary. Second, the algorithmic formulation of Network Dissection for pairing units and concepts assumes that concepts are all localizable in the image. The pairing is based on an intersection over union (IoU) criterion on the area of the image that produces the strongest activation of a unit and the segmentation of each concept in the Broden visual dictionary. In contrast, there are different facial concepts that simultaneously occur in the same region of the image, like ``mouth'' (a facial part), ``smiling'' (a facial attribute), ``wearing lipstick'' (a facial attribute), and ''AU-20'' (the action unit corresponding to lip stretcher). We refer to this challenge as the spatial overlap of concepts. Another important challenge is that we are interested in concepts that are not located in a specific region of the face, such as the apparent gender or the skin tone. We refer to this challenge as the global concepts challenge. In this case, the IoU-based nature of Network Dissection cannot be applied for quantifying global concepts. As described in \secref{interpretability}, the proposed Hierarchical Network Dissection is designed to deal with the spatial overlap of concepts and the global concepts challenges. 
We use Hierarchical Network Dissection to interpret and compare the representation learned by different face-centric models. Concretely, we dissect different deep learning architectures trained on popular publicly available facial datasets. These dissection results are presented and discussed in \secref{exp_real}. In particular, we can make very interesting observations. For example, the dissection reports revealed a gender bias in the training data of the Smile classification task, as well as the relevance of facial expression for the Beauty classifier. 

In the last part of our paper we conduct controlled experiments on biased datasets to showcase the potential of Hierarchical Network Dissection for bias discovery. Concretely, we consider the task of apparent gender recognition\footnote{We work with datasets with the binary labels corresponding to the social construct of gender classification. However we acknowledge that in essence the concept of gender is more fluid and non-binary.} and create biased training sets for specific concepts in our Face Dictionary. 
The dissection results show how our interpretability pipeline is able of revealing these biases, showing a significant amount of units that are paired to the concepts associated with the bias.

Both our Face Dictionary and the Hierarchical Network Dissection code for face-centric inference models are publicly released. 

\section{Related Work}
\label{sec:related}

\textbf{Model Interpretability} -- There are different techniques to understand the representations learned by deep convolutional neural networks (CNNs). Broadly, these techniques are based on: (\emph{i}) variants of backpropagation to visualize salient patterns, features, or image regions \cite{zeiler2014visualizing,simonyan2014deep, mahendran2015understanding}; (\emph{ii}) detecting patches in the image that strongly activate the units of the trained model \cite{object_detectors_emerge,zeiler2014visualizing}; or (\emph{iii}) analyzing to what extent units behave as detectors or classifiers for specific concepts 
\cite{object_detectors_emerge, network_dissection_cvpr} or synthesized ones \cite{nguyen2016synthesizing}. In this last direction, one of the most popular approaches is Network Dissection, proposed by \cite{network_dissection_cvpr, network_dissection_pami}. Network Dissection can be formulated to quantitatively analyze the representations learned by both classification and generative models \cite{gan_dissection, role_units_pnas}, making Network Dissection a very versatile approach. While Network Dissection has been widely used on object-centric and scene-centric models, the use of Network Dissection on face-centric inference models has remained unexplored. Our Hierarchical Network Dissection is inspired by Network Dissection.

\figDictionary


\textbf{Network Dissection} -- As first introduced in \cite{network_dissection_cvpr}, Network Dissection is a general framework for quantifying the interpretability of the hidden units in any convolutional layer of a neural network trained for an object-centric or a scene-centric task. It requires a broad range of visual concepts in order to compare the activation maps of these hidden units to their binary segmentation labels and compute their corresponding spatial affinity. This collection of concepts is presented as a visual dictionary called Broden, and it contains concepts that range from lower level (e.g. colors, textures, or materials, like red, dotted or metal) to high level (e.g. parts of objects, objects, or scenes, like leg, wheel, floor, car, or swimming pool). The pipeline uses these concepts and their binary segmentation masks to evaluate against the activation maps of a given layer in the network to generate an IoU score for each unit-concept pair. For each unit the concept with the highest IoU is reported and if the highest IoU is less than 0.04, the unit is called uninterpretable.

Notice that, as discussed in \secref{intro}, Network Dissection cannot be directly used to dissect face inference models. First, we need a face-centric dictionary instead of Broden. Second, we need to reformulate the unit-concept pairing algorithm to deal with the spatial overlap of concepts and the global concepts challenges. 

\textbf{Interpretability vs. Explainability} -- Model interpretability is related to the explainable AI topic \cite{arrieta2020explainable, gunning2017explainable}. There have been recent interesting efforts in explainable subject recognition systems \cite{yin2019towards,zee2019enhancing, explainable_face_rec_2020, richardwebster2018visual} and also in explainable face-centric generative adversarial networks (GANs) \cite{Shen2020}. Although Explainable AI and model interpretability are related topics, there are fundamental differences in the two concepts. While explainability focuses on generating explanations about the output, model interpretability focuses on understanding the internal representation of the model. Our work is centered around the latter topic by revealing the interpretable representation of face-centeric inference models.

\textbf{Bias Discovery} -- In the context of fairness in  artificial intelligence (AI), the interest for revealing bias in the data has gained a lot of attention among the research community. Wang et al. \cite{wang2020vibe} recently presented a detailed study and a computational tool to discover biases in image datasets. Their tool revealed very interesting insights on gender-based representation biases. Nonetheless, their approach is in general scene-centric and the tool is currently not targeted to be used in face-centric datasets. At the same time, \cite{wang2020towards} recently presented a benchmark to compare bias mitigation methods. The benchmark is created by introducing bias in an object classification dataset. In particular, per each class, a specific percentage of images are converted to grey-scale, creating an unbalanced representation (i.e. some classes have most of the images in grey-scale, while others have most of the images in color). The design of this benchmark has inspired our controlled experiments presented in \secref{exp_biased}. Furthermore, we demonstrate how model interpretability can be used as a tool for bias discovery. 
\section{Hierarchical Network Dissection}
\label{sec:dissection}

The unit-concept pairing of Hierarchical Network Dissection is based on our Face Dictionary, which is composed of several facial concepts. The concepts of the dictionary are organized in two main categories: Global Concepts and Local Concepts. The Local concepts are grouped by Facial Part and are organized in two categories: Action Units and Facial Attributes. Following this hierarchy of concepts in our Face Dictionary, Hierarchical Network Dissection uses three stages to determine what units in a CNN model are acting as concept detectors.

\subsection{Face Dictionary}
\label{sec:dictionary}



Our Face Dictionary consists of 12 Global concepts and 38 Local concepts, along with a collection of corresponding sample images per each concept. The images are collected from the following public face datasets: EmotioNet \cite{7780969}, UTKFace \cite{zhifei2017cvpr}, and CelebA \cite{liu2015faceattributes}. \figref{fig_dictionary}.a lists all the concepts included in the Face Dictionary. 

We have 4 types of Global concepts: apparent Age (4 categories), apparent Gender (2 categories), Ethnicity (4 categories), and Skin tone (2 categories)\footnote{We work with academic datasets with categorical labels corresponding to social constructs, apparent presence, or stereotypical representations of these concepts. However we acknowledge that these concepts are more fluid in essence, as further discussed in Sect. \ref{sec:exp_real}}. The total number of images corresponding to Global concepts in our dictionary is 11000, where apparent Age, apparent Gender, and Ethnicity have 4000 images each, and Skin Tone has a total of 3000 images. 

Regarding the Local concepts, we have two types: Action Units (denoted by AU), which refer to specific facial muscle movements; and facial attributes, referring to the permanent presence of a concept on the face. The local concepts are grouped according to the the facial region where they occur, as illustrated in \figref{fig_dictionary}.a. For the Local concepts our dictionary also includes segmentation masks that have been automatically estimated as we described below. \figref{fig_dictionary}.b shows some examples for local concepts and their corresponding segmentation masks. In total there are 24632 images corresponding to the local concepts. The histogram of the number of labelled instances per concept is displayed in \figref{dictionary_breakup}.




\DictionaryHist

\subsubsection{Automatic estimation of segmentations masks for the Local concepts}
There are several datasets that provide labels for action units and facial attributes merely indicating whether a particular attribute or action unit is present in a given image, without providing the corresponding location in the image. However, our dictionary needs the segmentation of the local concepts to be used during the concept-unit pairing. To automatically estimate segmentation masks for the local concepts we run a landmark detection algorithm (\cite{6909637}) on the images as we use the landmarks to estimate the face region where our local concepts lie. We estimate the center and covariance matrix of a 2D Gaussian confidence ellipse around a particular concept and generate a binary mask for each concept per image.



\subsection{Unit-Concept Pairing}
\label{sec:interpretability}
Our unit-concept pairing formulation has three stages. The first stage pairs units with Global concepts. Then, the second stage pairs units with Facial Parts. Finally, the third stage pairs those units that have been paired with a specific Facial Part with Attributes or Action Units that have spacial overlap with the corresponding Facial Part.

\textbf{Stage I: Global Concepts} -- This first stage is based on the idea that Global concepts belonging to the same category are mutually exclusive (for example, we assume that the apparent age of a face can not be [0-20] and [20-40] at the same time). Here it is important to state that each of these concepts is not restricted to the subgroups we have in our dictionary. They are perceived to be continuous in the real world such as mixed ethnicities, fluid gender, etc. In order to work within the bounds of our dictionary, we must assume such exclusivity to present a preliminary argument in the field of interpretability for such concepts. To pair a unit with a Global concept in one of the categories (e.g. Apparent Age), we
take a forward pass across all $N$ images in our dictionary for each global concept in the corresponding category and record the feature maps of the layer being dissected, while retaining the information about which map belongs to which subgroup. In order to compare the activations from these feature maps we assign a rank to each map based on their maximum activation score, where the map with the lowest score has rank $1$ and the highest score has rank $N$. Then, we initialize a score for each subgroup as $0$ and increment them as we iterate through all the maps by:
\begin{equation}
\vspace{-.3cm}
\label{eqn:1}
  \mathcal{CS}_{s} \pluseq (\mathcal{R}_{n} \times \mathcal{MS}_{n}),
\end{equation}
where $\mathcal{CS}_{s}$ is the concept score for subgroup $s$, $\mathcal{R}_{n}$ is the rank and $\mathcal{MS}_{n}$ is the maximum activation score of the $n^{th}$ map and each score is only incremented by maps that belong to $s$. We use this formulation to establish a pecking order among the different subgroups by accounting for the strength of their activations relative to each other. We then normalize each score by:

\ProbViz

\begin{equation}
\vspace{-.2cm}
\label{eqn:2}
  \mathcal{CS}_{s} = \mathcal{CS}_{s}/N_{s},
\end{equation}
where $N_{s}$ is the number of images that belong to $s$, to make the scores comparable. Finally, we divide each score by the sum of all scores to obtain a set of relative probabilities such that $ \sum_{s\in G} P_{s} = 1,$ where $P_{s}$ is the probability of $s$ and $G$ is the global concept being analyzed. The range of $\mathcal{R}$ remains constant for each individual analysis. Normalizing $\mathcal{R}$ itself is irrelevant since after Eq.1 and Eq.2, we normalize each concept score by the sum of all scores to obtain relative probabilities. Hence, only the maximum activation scores of the feature maps need to be normalized since each model produces activations in a different range. Using these probabilities we classify a unit as biased towards $s$ if $P_{s} > 0.3$ when $G$ is ``Age'' or ``Ethnicity'' and $P_{s} > 0.55$ when $G$ is ``Gender'' or ``Skin Tone''. For each concept, we choose the threshold to be 0.05 greater than the mean probability of the concept, which is 0.25 for Age and Ethnicity and 0.5 for Gender and Skin Tone. These thresholds are not as stringent because each score is only scaled by their respective rank and the cumulative scores of each subgroup are almost never too far away from each other. In \figref{ProbViz}.b, we visualize the average ratio of probability of units to the mean probability of their concept in the event that they exceed the mean probability. We can see that for all of the tasks the ratio crosses 1 by a very small margin, which shows how delicate our measurement for bias needs to be. Hence, we use lenient thresholds to detect units biased towards any subgroup.



\textbf{Stage II: Facial Parts} -- This stage is focused on pairing units with Facial Part concepts. Similar to the Stage I, we take a forward pass across all the images from the local set in our dictionary to store the activation maps of the layer being dissected and run network dissection as described in \cite{network_dissection_cvpr} to generate IoU scores for each local concept per unit. Since a unit generally produces strong IoU scores for multiple concepts within the same region, it is highly misleading to report a single concept with the highest IoU as interpretable. This is why we identify the region of the face that the concept with the highest IoU belongs to, and evaluate the relevance of every other concept in this region (as shown in \figref{fig_dictionary}) to the unit by establishing a probabilistic hierarchy amongst them using the same formulation introduced above with a minor variation.

\textbf{Stage III: Local Concepts} -- Per each unit that has been paired with a Face Part in Stage II, we extract all the activation maps that have labelled instances of at least one of the concepts belonging to this region. Then, we use the same steps shown in \eqnref{1} and \eqnref{2} but this time instead of subgroups, we have clustered concepts and due to the simultaneous presence of these concepts, one activation map can contribute to the concept score of more than one concept. Hence, this overlap of images among concepts can lead to scores that are not clearly distinguishable, which is why we add one more step to demarcate these concepts. We scale these concept scores by their respective IoUs estimated during Stage II by:
\begin{equation}
\vspace{-.2cm}
\label{eqn:4}
  \mathcal{CS}_{k} = \mathcal{CS}_{k} \times IoU_{k},
\end{equation}
where $k$ is one of the concepts in the region. The unit's ability to localize must play a significant role in determining interpretability as we learned from \cite{object_detectors_emerge} and \cite{network_dissection_cvpr}.

Then, we replicate the formulation $\sum_{s\in G} P_{s} = 1,$ and obtain relative probabilities per concept. If any concept's probability crosses a threshold of $(1.5/K)$, where $K$ is the total number of concepts in the region identified, we deem a unit to be interpretable for this concept. In contrast with global thresholds, here we use a threshold of 1.5 times the mean probability of a certain facial region because each concept score is scaled by their respective IoU estimated during Stage II. Because of this additional step each score is scaled strongly which establishes a hierarchy that is a lot more distinct compared to global concepts. This can be seen in \figref{ProbViz}.a, where we show the average probability ratio for each facial region on all the tasks. Here we observe that there is more variation in the probabilities and a lot of them fall in and around the 1.5 mark. The missing bars in the plot indicate no units were found in that region for the respective model. Thus choosing this threshold allows a unit to be interpretable for more than one concept by accounting for all aspects of affinity among the unit-concept pair.

\tblCNNmodelnew
\section{Dissecting General Face Inference Models}
\label{sec:exp_real}

We used the described Hierarchical Network Dissection pipeline to dissect models trained for five common face-centric tasks: age estimation, gender classification, beauty estimation, facial recognition, and smile classification. When certain facial tasks are implemented using deep learning models, there is often a gap in the definition of how these concepts are defined in the real world to the academic setting. For example, in today's world gender is a fluid concept and has several subgroups that belong to it, however in our case we study gender as a binary construct due to dearth of labelled data for others. Similarly, beauty estimation is highly controversial since it is an entirely subjective attribute that is correlated with multiple attributes, but we study beauty estimation through the lens of manually annotated labels that are provided via an aggregation of scores from multiple volunteers and try to study the model's representation within such context. It is important to state that our analysis is based on these assumptions and we attempt to study the representations of these models independent of such real world perspectives.
We try to use the same model architectures for most tasks in order to contrast the representations learned by similar architectures for different tasks. 

\subsection{Models and Datasets}
\label{subsec:models, datasets}

The face-centric tasks chosen for our experiments are age estimation, gender classification, beauty estimation, facial recognition and smile classification. \tabref{CNNmodel} summarizes the details of the dissected models, specifies the layer that we dissect in our experiments, and shows the performance obtained per model. Notice that we focus our interpretability experiments in the deeper layers of the models, since the concepts in our Face Dictionary are high level concepts. According to the previous studies, the high level concepts are mainly found in the layers that are closer to the output \cite{object_detectors_emerge, network_dissection_cvpr}. Notice also that for facial recognition we compare two models: one trained on FaceNet \cite{DBLP:journals/corr/SchroffKP15} and the other one train on FairFace \cite{DBLP:journals/corr/abs-1908-04913}. 

We train both the age and gender models on IMDB-WIKI dataset \cite{Rothe-IJCV-2018} with a backbone of ResNet-50. The IMDB-WIKI dataset consists of half a million celebrity images crawled from IMDB and Wikipedia with age and gender labels. Even though the dataset provides age estimation as a classification problem, we choose to train the model as a regression task to avoid parsing through mislabelled data as most of the labels are provided through date of birth and image timestamps crawled from the web. This may lead to bad accuracy in classification, whereas minor errors in labelling will be better handled by regression. Hence, we preprocess the data to remove invalid entries of age as well as remove corrupt images. After training for 2 epochs, the model converges with a mean absolute error (MAE) of 6.6015 on the test set. For the gender model, it also converges within 2 epochs to give a validation accuracy of 90.88\% on the test set.

For the beauty estimation problem we use SCUT-FBP5500 \cite{liang2018scut}, which is a diverse benchmark for multi-paradigm facial beauty prediction. It consists of 5500 images with a beauty score in the range of $[1-5]$. The dataset provides a train-test split of 60/40 with 5 fold cross validation. The authors of this paper provide pretrained models and we have dissected the ResNet-18 architecture that has a mean squared error of 0.137 on the validation set.


For the facial recognition task we dissect two models. One is FaceNet, that is an InceptionResNetV1 pre-trained on VGGface2, and the other is FairFace, a ResNet-50 we trained on the FairFace dataset. The pre-trained model has a validation accuracy of 99.6\% on LFW \cite{LFWTech}, while the ResNet-50 has been trained on FairFace with a validation accuracy of 86.7\%. We could not replicate the extremely high performance of FaceNet model because we could not use the necessary high batch sizes due to hardware restrictions and also because the dataset is smaller. FairFace only has about 100K images which we have used to generate 80,000 triplets per epoch. Notice, however, that the obtained accuracies are competitive and good enough 
to extract meaningful observations on the model interpretability. Recent work such as \cite{DBLP:journals/corr/abs-1801-07698} and \cite{DBLP:journals/corr/LiuWYLRS17} have attained accuracies greater than 99\% for facial verification and identification on massive datasets with millions of images such as MSCeleb-1M \cite{DBLP:journals/corr/GuoZHHG16} and VGGface2 \cite{DBLP:journals/corr/abs-1710-08092}.

Finally, we train a smile classification model with ResNet-50 on the subset of the CelebA dataset that has ``Smiling'' attribute labels. We ensure that none of the images in this training set are a part of our dictionary to avoid biased dissection. The training and validation sets have 6000 and 1200 images, respectively. After 4 training epochs, the model achieves an accuracy of 91.2\% on the validation data.

\subsection{Results and Discussion}

\figref{dissecting_face_models}.a shows the amount of interpretable units found in each model in terms of facial parts, which correspond to the output of Stage II of Hierarchical Network Dissection. We observe that, in general, the region that gets more units is the Cheek region for all the models except for the Beauty classifier, that has more units on the eye region. \figref{dissecting_face_models}.b shows the distribution on the types of concepts. We observe that attributes are the most common in all the models with two exceptions: Beauty and FairFace. For Beauty, the most represented type of concept is action units. One explanation might be that facial expressions play an important role in facial attractiveness perception \cite{tatarunaite2005facial}. For the case of FairFace, we observe that the most represented type of concept is facial parts. Some qualitative examples of detected local concepts per each model are given in \figref{ActivationResults}.

\dissectingfacemodels

\figref{NewDissectionResults} shows the interpretability reports for all the dissected models, both for our Hierarchical Network Dissection formulation (left) and the original Network Dissection formulation (right) with the Local concepts. By comparing the two dissection reports per each model, we can see the interpretability results of Hierarchical Network Dissection are more complete compared to the results obtained by Network Dissection: (\emph{i}) There is a clear increase in the sheer number of interpretable units that we find per concept using our approach since we do not restrict a unit's ability to interpret multiple concepts; (\emph{ii}) We observe that some of the concepts that are not revealed by Network Dissection are actually detected through our hierarchy, which shows a more diverse distribution of interpretable concepts; (\emph{iii}) In contrast to Network Dissection, our hierarchical approach can also pair units with global concepts. Furthermore, our approach allows us to determine how many of the units that are interpretable for a local concept are also interpretable for a global concept, allowing us to create richer visualizations with the overlap of local and global concepts. If we analyze the similarities of Hierarchical Network Dissection and Network Dissection results, we see that both approaches on average are prone to detect more concepts that are spatially dominant such as Cheek Region concepts and very few concepts in the Nose Region as a result of varying IoU scores. 

\HDissResults

\ActivationResults

\figCoverageComb

While the results of \figref{NewDissectionResults} show the representations learned by the models, we noticed that, in turn, they also reveal certain biases of the data the models were trained on. 
We can see in the case of age model, most of the units are biased towards different age groups. This is to be expected since the task itself relies on the model's ability to discriminate within different age groups and this is why most of the local concepts detected also overlap with age biased units. For the gender model we observe something very similar. We see that for the global concepts, the model mainly focuses on both male and female concepts specifically in a balanced manner. This again shows that the layer is correlating to the concepts that are discriminative for the task the model is trying to learn and how the formulation of Hierarchical Network Dissection is able to reveal this. 

In case of the beauty model we also observe that, out of all our models, it is the only one that has a significant amount of units paired with facial Action Units. This points to the relevance of these concepts to the subjective nature of beauty and the influence of facial expressions on beauty perception, as argued before. Also, we see that the global concepts that is most represented in the unit-concept pairing is related to the skin tone type. This might be associated to the fact that the appearance of the skin is also a relevant factor for the stereotypical judgments of beauty. For example, in some cultures a tanned skin is considered more beautiful than an untanned one \cite{xie2013white}. 

For the facial recognition task our two models have been trained on datasets of vastly different size and demographic parity. We observe that FaceNet, which has been trained on millions of images, has an extremely diverse representation of concepts. Among the dissected models, it is the only one that captures a vast majority of local concepts and has an extremely high bias across all 4 global concepts. This may point to the fact that such models, when trained on massive datasets, may encode the inherent biases stemming from the distribution of these global concepts in the dataset. However the Fairface model is very selective and has only shown to focus on a very restricted set of concepts specially for the local concepts. Most of the local concepts found during the concept unit pairing are spatially dominant in terms of their area, which may also suggest that the interpretation reveals concepts that are more salient than the targeted task. Also for each global concept it has shown a significantly higher affinity towards individual subgroups rather than a concept as a whole. This points towards the model's inability to generalize well for different subgroups and hints that the representation learned by this model is poorer than its pretrained counterpart.

Finally, in the case of Smile model we observe something unexpected. We see that most units have been paired with a specific gender. This affinity is also supported by the local concepts, since most units have a preference towards No Beard and Rosy Cheek, two attributes that are mostly present in female faces. This is surprising since there should be no correlation between smile and the gender of a person. This forced us to investigate our training set which then revealed that there was a clear skew of 60\% females vs 40\% males in the training set for the smile class. The results suggest that the bias was detecting during the training and it was used for the model to converge quicker.

From a qualitative perspective, \figref{ActivationResults} shows examples of 2 units per model where our formulation generates a higher probability score for a concept which has a lower IoU than the concept identified by Stage II. We display the concept chosen purely by IoU at the top and the concept with the highest probability from our formulation at the bottom. As stated earlier, when applying the IoU approach from the original Network Dissection on the images from our dictionary it is impossible to say that a concept that generates a similar IoU to the top concept cannot be reported as an interpretable concept without doing a deeper analysis of the activations generated for that concept with respect to every single concept that lies in the same facial region. Thus, by establishing a hierarchy among concepts from that facial region, we learn that one unit can be paired with more than one concept.
This behavior can be explained by the spatial proximity that these concepts have and how difficult it is for a unit to distinguish between concepts that look almost the same but have different characteristics. 
For example, we see this happens for the gender model. Concretely, in Unit 142 of the second last convolution layer of ResNet-50 we see that the original approach identifies ``5 o Clock Shadow'' as the top concept with a very high IoU score of 0.1914. By observing the images we can see that the unit does a very good job in localizing the region with this concept. However when we use our formulation to estimate probabilities for all the concepts in the Cheek Region, we learn that ``Rosy Cheeks'' despite having a far lower IoU score of 0.1109 generates a probability of 0.2276 which is much higher than the 0.1884 generated by ``5 o Clock Shadow''. This illustrates our idea of ignoring concepts through the IoU approach since it would lead to an incomplete understanding of the model's representations.

\subsection{Interpretability Coverage Report}

Our current dictionary has a combination of local and global concepts and Hierarchical Network Dissection pairs individual units of a layer with them. However, it is important to notice that there are many facial concepts that are not included in the dictionary and this limits the interpretability capacity of Hierarchical Network Dissection. This section presents a quantitative analysis on the amount of units that Hierarchical Netork Dissection, along with our Face Dictionary, is able to successfully pair with facial concepts.
For this goal we have dissected the deeper layers of all the six tasks discussed above using our hierarchical formulation as well as the original Network Dissection. 

We display in Fig. \ref{fig_coverage_comb}
the number of interpretable units per layer for the different trained models (one per row), and for both the Hierarchical Netowk Dissection (left) and Network Dissection (right). For the models in Table \ref{tbl:CNNmodel} with ResNet as their backbone we have dissected all the convolutional layers in Block3 and Block4 (notice that ResNet-50 and ResNet-18 have 4 blocks of convolutional layers). For the FaceNet model, we dissect the final convolution block as well as all the convolution layers of the 5 Inception-C blocks (which are named as repeat3 in the figures). The reason we avoid dissecting the earlier layers of the network is due to the extremely high resolution of the feature maps at this stage of the network, as well as a model's inability to localize higher level concepts which was shown by \cite{network_dissection_cvpr}. 

We can observe that, except the FaceNet model which has very rich representations capable of detecting many concepts at several stages of the network, most of the other networks show a roughly 70-80\% coverage across multiple layers through the hierarchical approach (Fig. \ref{fig_coverage_comb}.b, left) and around 50-60\% coverage using the network dissection approach (Fig. \ref{fig_coverage_comb}.b, right). Surprisingly, we do not observe a striking difference in the percentage of interpretable units across the upper and deeper layers, which suggests that the models have a relatively strong understanding of the underlying concepts even at the middle stages. 

We must point out that due to the inability of the original dissection to account for global concepts it is not strictly fair to compare these two approaches directly in terms of coverage. Our formulation allows us to account for a variety of concepts which helps us to create a more complete understanding of the model's representations and highlights it's superiority. This primarily ensures a high coverage across most models since a unit that cannot effectively localize certain concepts, may be susceptible to biases towards a subgroup of global concepts. This is easy to observe in Fig. \ref{fig_coverage_comb},
where Fairface and Beauty models show the lowest coverage across all layers with very few layers displaying a 50\% coverage. This coverage may be impacted by several factors, such as the distribution of correlated and uncorrelated attributes within a dataset and the training process of the model. This is why it is beneficial for the models to be exposed to as many concepts as possible in order to improve the estimation of these relationships derived from Hierarchical Network Dissection. A more diverse and rich version of the dictionary with several other concepts would allow us to push the limits of this formulation and create richer dissection reports for every single face inference model.



\section{Hierarchical Network Dissection for Bias Discovery}
\label{sec:exp_biased}




This section presents our experiments to showcase the potential of Hierarchical Network Dissection for bias discovery. Since our Face Dictionary includes two types of concepts (Global and Local), we perform one set of experiments per each type of concept. 

Concretely, we consider the gender classification task and we explicitly introduce different degrees of bias in the training set using first the local concept ``Eyeglasses'' and second the global concept ``Gray-scale''. Introducing controlled bias is a common strategy in bias mitigation (e.g. \cite{wang2020towards}). However, in this case we are not proposing a bias mitigation technique. We are showing how Hierarchical Network Dissection can provide insights regarding dimensions of potential bias that is represented in the model. 
Notice that revealing bias is important to improve the creation of datasets and also to apply bias mitigation methods that take the dimensions containing bias as input \cite{bahng2020learning}, \cite{tartaglione2021end}, \cite{clark2019dont}, assuming that it is already known where the bias is present.

\Simulatedglassesimages
\subsection{Bias Introduced on a Local Concept}
\label{sec:exp_biased_gender_local}



In our first experiment, we use the local concept ``Eyeglasses'' from our Face Dictionary to create six different biased datasets. Concretely, a percentage $P$ of males in the training set will be wearing eyeglasses, while a a percentage $(100-P)$ of females will be wearing eyeglasses. \figref{eyeglasses_bias} illustrates the training data for the two extreme cases: $P = 50$ (no bias introduced) and $P=100$ (maximum degree of bias, where detecting faces with eyeglasses is equivalent to do gender classification). Notice that the closer $P$ is to $100$ the more useful would be for the model to have an internal representation that focuses more in the eye region, since concept ``Eyeglasses'' belongs to eye region and it can help to discriminate on the main task (i.e. gender classification). 

\TopGlassesImages
\SimulatedExp

We train with a ResNet-50 architecture and dissect the second to last convolution layer for each case. When we observed the number of interpretable units for ``Eyeglasses'' in particular, only the models with $P=50$ (balanced) and $P=100$ (fully biased) showed a contrast, where the fully biased model detects more than thrice the number of units as the balanced model whereas the models in between detected a similar number of units closer to the balanced model. In \figref{results_eyeglasses_bias}, we show per each degree of bias the number of interpretable units in terms of facial parts. Notice that the number of units that focus on the eye region increases as $P$ increases. This result reveals how the representation of the model focuses more in eye region as the discriminant information contributed by the eye region increases. Interestingly, we also observe that the focus on the cheek region decreases as $P$ increases. This again hints to the fact that hidden units may only be exclusively discriminative in terms of facial region and not so much for individual facial concepts.

In \figref{top_gender_glasses}, we show a set of qualitative results to visualize the top 50 activated images per gender - female (left), male (right) - for each model. We observe that the distribution of images with ``Eyeglasses'' goes up for males as P value increases from 50 to 100 and the exact opposite can be seen for the female class as the number of images with ``Eyeglasses'' steadily decreases. This shows that the model's learning becomes increasingly biased as the bias in the dataset goes up. The images are arranged in descending order based on their maximum activation scores.

\subsection{Bias Introduced on a Global Concept}
\label{sec:exp_biased_gender_non-local}

Our second experiment follows a similar protocol as the previous experiment, but the bias is introduced on the global concept ``Grayscale''. Thus, our training datasets will have an unbalanced representation of color images and grey-scale images: for males, a percentage $P$ of the images will be gray-sale, and for females a percentage $100-P$ of the images will be gray-scale. 
We synthesize 6 different datasets and train gender classification models, with a percentage $P$ bias going from 50 to 100. 

\grayscalePROBS

We use Stage I of Hierarchical Network Dissection by taking a separate test set to generate unit probabilities for all the units in the second to last convolution layer of block \#1, \#2, \#3, and \#4 of the ResNet-50 architecture. For any given unit $u$, probability of color $P_{c}^{u}$ and probability of gray-scale $P_{g}^{u}$ represent the affinity of the unit to a color scheme, where $P_{c}^{u} + P_{g}^{u} = 1$. We perform this at all stages of the network because color scheme is a low-level concept and can be detected in earlier stages of the network. We compare these unit probabilities across different models for all 4 layers with $64, 128, 256$ and $512$ convolution filters respectively to interpret how the bias within each layer increases or decrease as the bias rises from $P=50$ to $P=100$.

In \figref{grayscale_probs}, we display the concept probabilities ($P_{color}$ \& $P_{gray}$) for all the units in an ascending manner to show the gradual shift in the number of biased units. If the majority of units in a layer are not biased, we should get a very gentle slope and a highly biased group of units should return a steep slope. In this case, we clearly observe that the balanced set with P=50 displays a gentle slope across all 4 layers and as the P value increases, the slope of probabilities gradually becomes more steep. We observe that in the case of P=100 (even in the early layers), there are very few units that do not have a strong affinity to one of the color schemes verifying that our formulation can easily classify units within a layer to hypothesize how biased their representations are.
\nonlocalizable

For each layer, we also establish the number of biased units to compare how biased the representations are at different stages of the network. A unit is said to be biased if either $P_{c} > 0.55$ or $P_{g} > 0.55$ staying consistent with the bias analysis introduced in the paper. In \figref{non_localizable_results}, we observe that all 4 layers show a clear increase in the number of biased units as we move from the unbiased set to the completely biased one. Apart from Layer 3 that has shows a perfectly uniform increase, the other layers display fluctuation in the level of bias but eventually showcase a steep increase as the $P$ values reaches its maximum value. We also observe that the last layer has a far greater percentage of biased units across all $P$ values, which is reasonable when we think about how crucial this perceived bias is to the model's performance due to its proximity to the final fully connected layer. This experiment in tandem with \secref{exp_biased_gender_local} emphasizes the ability of our formulation to discover and quantify the biases that exist within the representations, which are usually hidden from us due to our inability to interpret these models quantitatively.

\section{Conclusions}
In this paper, we present a general pipeline to interpret the internal representation of face-centric inference models called Hierarchical Network Dissection. Our approach is inspired by the Network Dissection work, which is a well-known object-centric and scene-centric model interpretability pipeline. The proposed Hierarchical Network Dissection formulation can deal with two challenges of face-centric model interpretability: ``spatial overlap of concepts'' and ``global concepts''. In summary, the main contributions of our work are: (1) we created our ``Face Dictionary'', a collection of face-centric concepts with corresponding image samples; (2) we introduced Hierarchical Network Dissection, an algorithmic approach that uses a probabilistic formulation to pair units with global and local; (3) we create interpretability reports for different face-centric inference models that have been trained on popular facial datasets and tasks, and provide an extended discussion on how the model interpretability can be used to better understand how the model works and how the model represents the information; and (4) we performed controlled experiments on biased data to empirically show how the dissection of the model can be used for bias discovery. 
Our ``Face Dictionary'' and the code of Hierarchical Network Dissection will be publicly released. We hope them to be useful tools for the computer vision community working on face inference models. 

\section*{Acknowledgments}
This work was partially supported by the Spanish Ministry of Science, Innovation and Universities, RTI2018-095232-B-C22. 

\bibliographystyle{spbasic}   
\bibliography{paper}  

\end{document}